\documentclass[runningheads,a4paper]{llncs}

\usepackage{amssymb}
\usepackage{graphicx}
\usepackage{algorithm2e}
\usepackage{url}

\urldef{\mailsa}\path|{felipe.elorrieta@usach.cl}|    
\newcommand{\keywords}[1]{\par\addvspace\baselineskip
\noindent\keywordname\enspace\ignorespaces#1}

\begin{document}

\mainmatter  

\title{Online estimation methods for irregular autoregressive models \thanks{Supported by ANID Fondecyt Iniciacion 11200590 grant and the ANID Millennium Science Initiative ICN12\_009, awarded to the Millennium Institute of Astrophysics.}}

\titlerunning{Online irregular autoregressive models}

%
%
\author{Felipe Elorrieta\inst{1,3}\orcidID{0000-0002-1835-7433}%
\and Lucas Osses \inst{1} \and Matias Cáceres \inst{1} \and Susana Eyheramendy\inst{2,3}\orcidID{0000-0003-4723-9660} \and
Wilfredo Palma\inst{3}}
\authorrunning{Elorrieta.F et al.}

\institute{Department of Mathematics and Computer Science, Universidad de Santiago, Chile\\
\mailsa\\
\and Faculty of Engineering and Sciences, Universidad Adolfo Iba\~nez, Chile\\
\and Millennium Institute of Astrophysics,  Santiago, Chile}

%
%

\maketitle

\begin{abstract}
In the last decades, due to the huge technological growth observed, it has become increasingly common that a collection of temporal data rapidly accumulates in vast amounts. This provides an opportunity for extracting valuable information through the estimation of increasingly precise models. But at the same time it imposes the challenge of continuously updating the models as new data become available.

Currently available methods for addressing this problem, the so-called online learning methods, use current parameter estimations and novel data to update the estimators. These approaches avoid using the full raw data and speeding up the computations.

In this work we consider three online learning algorithms for parameters estimation in the context of time series models. In particular, the methods implemented are: gradient descent, Newton-step and Kalman filter recursions. These algorithms are applied to the recently developed irregularly observed autoregressive (iAR) model. The estimation accuracy of the proposed methods is assessed by means of Monte Carlo experiments.

The results obtained show that the proposed online estimation methods allow for a precise  estimation of the parameters that generate the data both for the regularly and irregularly observed time series. These online approaches are numerically efficient, allowing substantial computational time savings. Moreover, we show that the proposed methods are able to adapt the parameter estimates quickly when the time series behavior changes, unlike batch estimation methods.

\end{abstract}

\section{Introduction}

Several natural or social phenomena can be measured sequentially over time. There are a wide range of tools to explain the temporal behavior of these phenomena. Among the most popular methods are the autoregressive moving average (ARMA) models. This model can be useful, for example, to explain the time dependence of a stationary time series and to forecast its future behavior. However, these forecasts become very uncertain over a large prediction horizon. This is an issue, since in many cases temporal data are still being collected, which makes the time series models useless if they are not updated recurrently.\\ 

This updating can be done in two ways: in batch or online setting. In the online setting each training instance is processed once “on arrival” without the need for storage and reprocessing. In contrast, for batch learning, the whole dataset is required to recalculate the current learning.\\

Avoiding storing and reprocessing previous information can be very useful to optimize the computation time of the models if we need to fit large time series or if we intend to model many time series simultaneously. The second case is a very common challenge in time series classification based on features that are extracted from the temporal behavior of each time series \cite{Maharaj_2019}. For example, in astronomy, astronomical objects can be classified by the temporal behavior of their brightness \cite{Debosscher_2007,Richards_2011,Elorrieta_2016,Sanchez_2021}. However, currently astronomical data are processed in real time \cite{Bellm_2018,Ivezic_2019} so that constant updating of time series models is a challenge which is addressed in this work.\\

Several works have already addressed the problem of online estimation in regularly observed time series \cite{Anava_2013,Liu_2016}. However, in some cases temporal data may not be obtained regularly. This problem occurs frequently in astronomy \cite{Feigelson_2018}, climatology \cite{Mudelsee_2014} or high-frequency finance \cite{Hautsch_2012}.\\

Some models have been proposed in the literature to fit irregularly observed time series. These models can be separated into two groups. The first one includes models that assume continuous times, such as the Continuous Autoregressive Moving Average (CARMA \cite{Jones_1981}) or Continuous-Time Fractionally Integrated ARMA Process (CARFIMA \cite{Tsai_2009}) models. These models are the solution of differential stochastic equations and assume that there are small time gaps between observations.\\

On the other hand, the second group considers discrete times representation of irregularly observed time series. Some models that follows this approach are the irregular autoregressive (iAR \cite{Eyheramendy_2018}), complex irregular autoregressive (CiAR \cite{Elorrieta_2019}), bivariate irregular autoregressive (BiAR \cite{Elorrieta_2021}) and the irregular autogressive moving average (iARMA \cite{Ojeda_2021}) models.\\

In this work, we are particularly interested in proposing online estimation methods for the irregularly observed autoregressive (iAR) model. For this purpose, we follow two different approaches. First, following the work of Anava et al. \cite{Anava_2013}, we use numerical methods such as Gradient Descent and Newton-Raphson to derive an update equation for the parameter of the iAR model. In the second approach, we obtain a recursive estimate of the a posterior distribution of the model parameter, under a Bayesian structure \cite{Saerkkae_2013}.\\

The structure of this paper is as follows. In Section \ref{sec:methods} we describe the irregularly observed autoregressive model and the three online estimation methods proposed in this work. In Section \ref{sec:simulation} we perform Monte Carlo experiments to evaluate the proposed online estimation methods in different simulated scenarios. The application to real-life data is shown in Section \ref{sec:application}. Finally, the main findings of this work are discussed in Section \ref{sec:discussion}.

\section{Methods}\label{sec:methods}

\subsection{irregular Autoregressive model}

The irregular Autoregressive (iAR) model was introduced by Eyheramendy et al. (2018)\cite{Eyheramendy_2018} and it is defined as:

\begin{equation} \label{ec2}
y_{t_j}=\phi^{t_j-t_j-1} y_{t_{j-1}} +\sigma\sqrt{1-\phi^{2(t_j-t_{j-1})}}\epsilon_{t_j},
\end{equation}
where $\epsilon_{t_j}$ is a white noise sequence with zero mean and unit variance, $\sigma$ is the standard deviation of $y_{t_j}$ and $t_j$ are the observational times such that $j=1, ..., n$.

The minus log-likelihood function of the process, under Gaussianity, is given by,

\begin{equation} \label{ec3}
l(\theta)=\frac{n}{2}\log(2\pi)+\frac{1}{2} \displaystyle\sum_{i=1}^{n}\log v_{t_j}+ \frac{1}{2} \displaystyle\sum_{i=1}^{n} \frac{e^2_{t_j}}{v_{t_j}},
\end{equation}
where the initial values are defined by $e_{t_1}=y_{t_j}$, $v_{t_1}=\sigma^2+\delta^2_{t_1}$. In addition, the one-step predictor is defined by $\hat{y}_{t_j}=\phi^{t_j-t_{j-1}}y_{t_{j-1}}$, while the innovation is defined as $e_{t_j}=y_{t_j}-\hat{y}_{t_j}$. The innovation variance is $v_{t_j}=\sigma^2 (1-\phi^{2(t_j-t_{j-1})})+\delta^2_{t_j}$, where $\delta^2_{t_j}$ is the known variance of the measurement errors. Note that the parameter $\phi$ describes the autocorrelation function of the process. Furthermore, $y_{t_j}$ is a weakly stationary process under $0<\phi<1$. \\

Generally, this model is fitted in an irregularly observed time series using all the information available, i.e., in a batch setting. As mentioned above, in this work we propose online estimation procedures for the iAR model. With these methods, the estimation of the parameter $\phi$ can be updated when new observations arrive without the need to reprocess the previous information. In this work we propose three online estimation methods, which are introduced in the following subsection.

\subsection{Online estimation algorithms}

A first approach to implement the online estimation algorithms for the iAR model is based on the first order optimization of a given loss function \cite{Anava_2013}. We propose two estimation algorithms which follow this idea. The first one, is called the Online Newton Step (ONS) method. The ONS method was introduced by Hazan et al. (2007) \cite{NS} and is based on the Newton-Raphson Method. The main goal of this method is to exploit the information to accelerate the convergence of the optimization. On the other hand, we propose to use the Online Gradient Descent (OGD) method (\cite{Cesa_1996},\cite{GD}), which aims to optimize the parameter estimation by computing the gradient of a loss function.\\

A second approach is based on a bayesian estimation method for the linear regression model. The algorithm that was built following this idea is the Online Bayesian Regression (OBR), which assumes a prior distribution on the parameter of the model and combines it with the likelihood of the observations in order to obtain the posterior distribution. The parameters of the resulting distribution contains the estimation results for the parameter of the iAR model.

\subsubsection{iAR Online Newton Step (iAR-ONS)}

The ONS algorithm uses the second derivative of the loss function. The adaptation of this algorithm for online estimation of the parameter $(\phi)$ of the iAR process is as follows, \newline

\begin{algorithm}[H]
 \caption{iAR-ONS}
     \SetAlgoLined
 Input: $\phi_{1}$ initial value; learning rate $\eta$; $A_1=\eta$.
 \BlankLine
\For{$j=1$ \KwTo n-1}{
 Get $\hat{y}_{t_j}=\phi_j^{t_j-t_{j-1}}y_{t_{j-1}}$\\
 Observe $y_{t_j}$ and obtain the loss function $l_{j}(\phi_{j})$;\\
 Let $\bigtriangledown_j=\bigtriangledown l_{j}(\phi_{j})$;
 Update $A_{j+1}=A_{j}+\bigtriangledown_j\bigtriangledown_j$;\\
 Parameter update $\phi_{j+1}=(\phi_{j}-\frac{1}{\eta}A_{j+1}^{-1}\bigtriangledown_j)$
}
\end{algorithm}

In our implementation, we assume that $l_{j}(\phi_{j})$ is the quadratic loss function defined by $(y_{t{j}}-\hat{y}_{t_j})^2$. Later, the gradient is defined by $\bigtriangledown_j = -2(y_{t{j}}-\hat{y}_{t_j})y_{t_{j-1}} (t_j-t_{j-1})\phi_j^{t_j-t_{j-1}-1}$

\subsubsection{iAR Online Gradient Descent (iAR-OGD)}

The second online estimation method that we propose in this work is the iAR-OGD algorithm, which aims to optimize the parameter estimation of the iAR process using the gradient descent method.The steps to implement this algorithm are the following,\\

\begin{algorithm}[H]

 Input: $\phi_{1}$ initial value; learning rate $\eta$.

\For{$j=1$ \KwTo n-1}{
 Get $\hat{y}_{t_j}=\phi_j^{t_j-t_{j-1}}y_{t_{j-1}}$;\\
 Observe $y_{t_j}$ and obtain the loss function $l_{j}(\phi_{j})$;\\
 Let $\bigtriangledown_j=\bigtriangledown l_{j}(\phi_{j})$;\\
 Parameter update $\phi_{j+1}=(\phi_{j}-\frac{1}{\eta}\bigtriangledown_j)$
}
 \caption{iAR-OGD}
 \end{algorithm}
 
Here the loss function $l_{j}(\phi_{j})$ and the gradient $\bigtriangledown_j$ are defined as in the iAR-ONS algorithm. 

\subsubsection{iAR Online Bayesian Regression (iAR-OBR)}

The OBR algorithm assumes a Gaussian prior distribution on the parameter $\phi$ of the iAR model, and combines it with the likelihood of the current observation being analyzed, which is also Gaussian. In addition, the parameter $\phi$ is defined by a random walk between the observations \cite{Saerkkae_2013}. This information can be combined to obtain the posterior distribution of the parameter of the iAR model given the first $j$ observations, in which its mean and its variance are the estimated parameter ($\phi_{j}$) and its variance ($P_{j}$) respectively. The parameters of the posterior distribution can be obtained using a modified version of the Kalman recursions, which are defined as follows,    \newline

\begin{algorithm}[H]
 \caption{iAR-OBR}
     \SetAlgoLined
 Input: $\phi_{1}$, $P_{1}$ initial values.
 \BlankLine
\For{$j=2$ \KwTo n}{
 Observe $y_{t_{j-1}}$ and $y_{t_j}$;\\
 Let $S_{j}= y_{t_{j-1}} \; P_{j-1} \; y_{t_{j-1}}\; +\; \sigma^2(1-\phi_{j-1}^{2({t_j- t_{j-1}})})$;\\
 and $K_{j} = P_{j-1}\; y_{t_{j-1}} \; S_j^{-1}$;\\
 Update $\phi_{j} = \phi_{j-1}\; +\; K_j \left[y_{t_{j}}\;-\; y_{t_{j-1}} \phi_{j-1}^{({t_j- t_{j-1}})}\right]$;\\
 and $P_{j}= P_{j-1}\; -\; K_j\; S_j\; K_j$;\\
}
\end{algorithm}



\section{Simulation Experiments}
\label{sec:simulation}

In this section we assess the parameter estimation accuracy of the three online estimation methods for the iAR model proposed in this work. For the following experiments, we generate each iAR process with 400 observations with random observational times coming from the following four distributions:

\begin{itemize}
    \item Regular time with constants gaps.
    \item Time gaps following a Uniform distribution with parameters $a=0.5$ and $b=1.5$.
    \item Time gaps following a Gamma distribution with parameters $\alpha=3$ and $\beta=3$.
    \item Time gaps following a mixture of two exponential distributions with means $\lambda_1=15$ and $\lambda_2=2$ and weights $w_1=0.15$ and $w_2=0.85$ respectively, such that, 
    
    $$f(\nabla_j|\lambda_1,\lambda_2,\omega_1,\omega_2)=\omega_1 g(\nabla_j|\lambda_1)+\omega_2g(\nabla_j|\lambda_2) ~~~ \forall j=2,\ldots,n$$
    
    where $g(.)$ is the exponential distribution and $\nabla_j = t_j - t_{j-1}$ is the j-th time gap. 
\end{itemize}

We chose these distributions in order to assess whether the size of the gaps in which the observations are taken can affect the results of the estimation methods. Both for the generation of the iAR process as well as the observational times, we use functions available in the iAR package of R \cite{iAR}.\\ 

For each time distribution we implemented the three following scenarios. First, we perform a simple Sanity Check, in which we generate the iAR model with a parameter $\phi$ and assess whether the proposed methods estimate the parameter accurately. The second experiment consists in an Abrupt Change in the parameter of the model,i.e.,the parameter of the model remains constant for the first half of each simulated iAR process and for the second half it changes to another value. Finally, the last scenario is the Constant Change, in which the parameter of the model change slowly in time. The Abrupt Change and the Constant Change experiments allow to evaluate whether the online estimation can adapt to changes in the process structure.\\

Following the Monte Carlo method, each experiment was repeated 100 times. In order to assess the proposed estimation methods we use three evaluation measures. The first of them, is the parameter estimation accuracy. Second, we evaluate the goodness of fit by computing the Mean Squared Error (MSE) of the fitted values. Finally, we evaluate the computation time of the proposed algorithms.\\

For the Sanity Check and Constant Change scenarios we assumed that the first 50\% of the measurements of the time series were observed, while for the Abrupt Change scenario we assumed that the initial 62.5\% of the time series was observed. From the observed values we estimate the iAR model parameter in a batch setting. This batch estimate was used as the initial value of the parameter for the online estimation methods performed for the remaining observations. Finally, in each scenario we also compare the online estimation methods with the batch estimation which uses all the values of each time series. The batch estimation of the iAR model is obtained from the IARloglik function of the iAR package. From now we will refer to it as iAR-MLE.\\


Table \ref{sim1} shows a summary of the Monte Carlo experiments for the Sanity Check scenario where the parameter of the simulated model is $\phi = 0.5$ for the four distributions used to generate the observational times. In the table we present the last value estimated by each estimation method, denoted by $\hat{\phi}_{400}$. Note that the three online estimation methods achieve a fairly accurate estimate at the end of the time series. In the comparison between the online estimation methods, the iAR-ONS method stands out with a closer estimation to the true value of the parameter for three of the four time distributions.\\

In order to assess to the goodness-of-fit of each estimation method, we present the MSE obtained from the fitted values estimated from the beginning of the online estimation until the end of each simulated series. From the MSE we observe that the batch estimation method has higher values than the iAR-OBR method, indicating that the use of this online estimation method gives better results for the fitted values in this setting.

\begin{table}
\begin{centering}
\resizebox{\columnwidth}{!}{
\begin{tabular}{|c|c|c|c|c||c|c|c|c|}
  \hline
\textbf{Obs. Time} & \textbf{$\hat{\phi}_{400}$ MLE} & \textbf{$\hat{\phi}_{400}$ OBR} & \textbf{$\hat{\phi}_{400}$ OGD} & \textbf{$\hat{\phi}_{400}$ ONS} & \textbf{MSE MLE} & \textbf{MSE OBR} & \textbf{MSE OGD} & \textbf{MSE ONS} \\ \hline \hline
  Regular & 0.502 & 0.48 & 0.49 & \textbf{0.504} & 0.74 & \textbf{0.7} & 0.74 & 0.78 \\ 
  Unif(0.5,1.5) & 0.5 & 0.475 & 0.479 & \textbf{0.496} & 0.74 & \textbf{0.7} & 0.74 & 0.77 \\ 
  Gamma(3,3) & 0.502 & 0.473 & 0.484 & \textbf{0.494} & 0.66 & \textbf{0.62} & 0.66 & 0.69 \\ 
  Exp. M(15,2,0.15,0.85) & 0.503 & \textbf{0.481} & 0.481 & 0.533 & 0.76 & \textbf{0.71} & 0.73 & 0.8 \\ 
   \hline
\end{tabular}
}
\end{centering}
\caption{Summary of the Monte Carlo experiments for the Sanity Check scenario performed using a parameter $\phi=0.5$ in the iAR model for the four distributions used to generate the observational times. Columns 2 to 5 of this table show the last value estimated for each method. The last four columns show the mean squared error (MSE) computed for the fitted values.\label{sim1}} 
\end{table}

\begin{center}
\begin{figure*}
\begin{minipage}{0.325\linewidth}
\centering
\includegraphics[width=\textwidth]{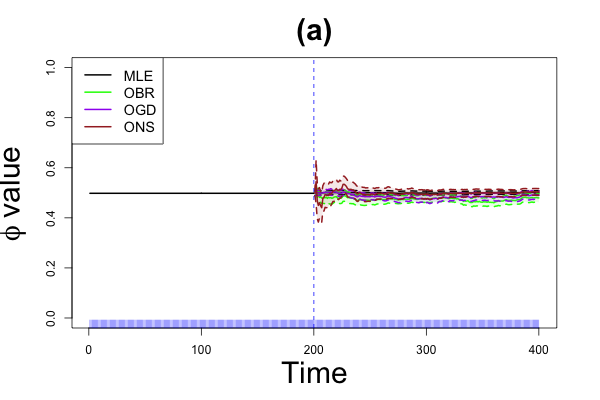}
\end{minipage}
\begin{minipage}{0.33\linewidth}
\centering
\includegraphics[width=\textwidth]{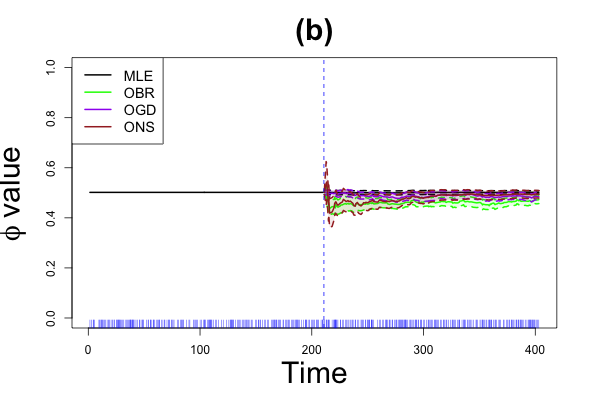}
\end{minipage}
\begin{minipage}{0.33\linewidth}
\includegraphics[width=\textwidth]{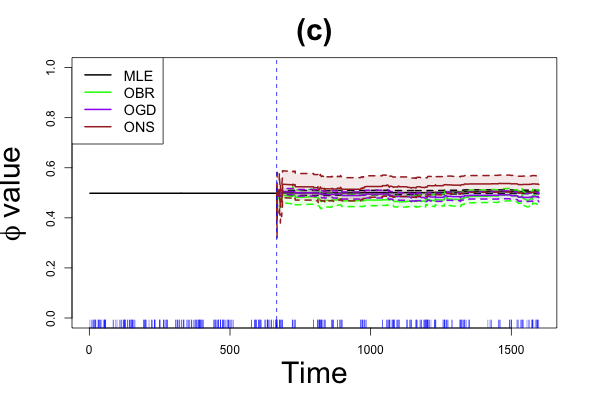}
\end{minipage}
\begin{minipage}{0.325\linewidth}
\centering
\includegraphics[width=\textwidth]{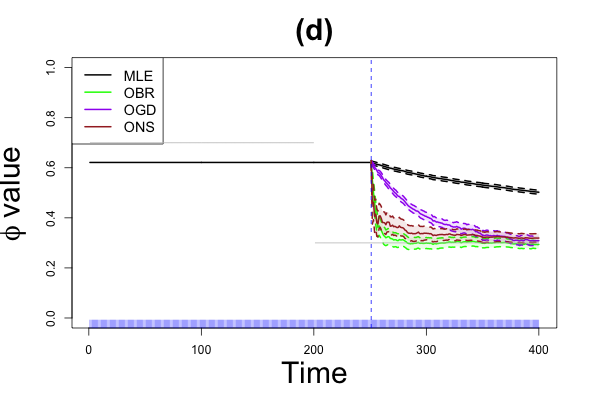}
\end{minipage}
\begin{minipage}{0.33\linewidth}
\centering
\includegraphics[width=\textwidth]{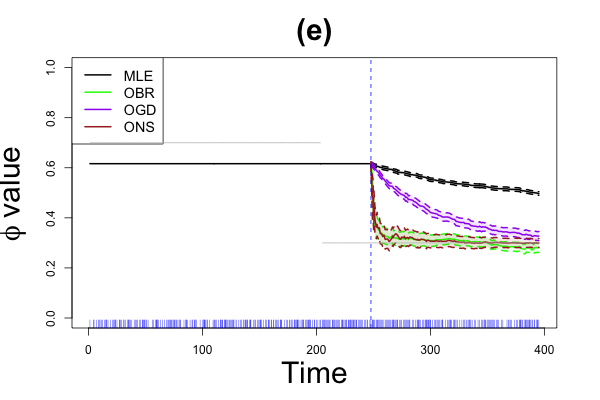}
\end{minipage}
\begin{minipage}{0.33\linewidth}
\includegraphics[width=\textwidth]{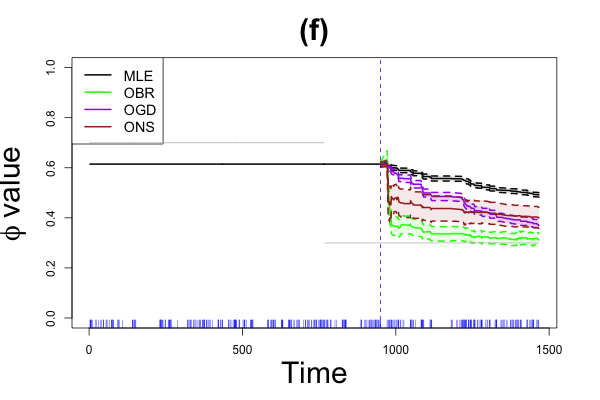}
\end{minipage}
\begin{minipage}{0.325\linewidth}
\centering
\includegraphics[width=\textwidth]{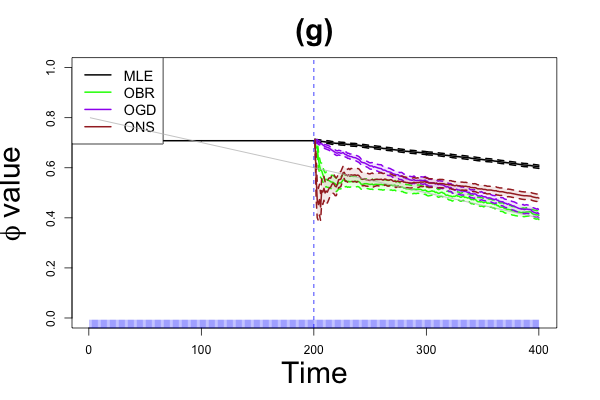}
\end{minipage}
\begin{minipage}{0.33\linewidth}
\centering
\includegraphics[width=\textwidth]{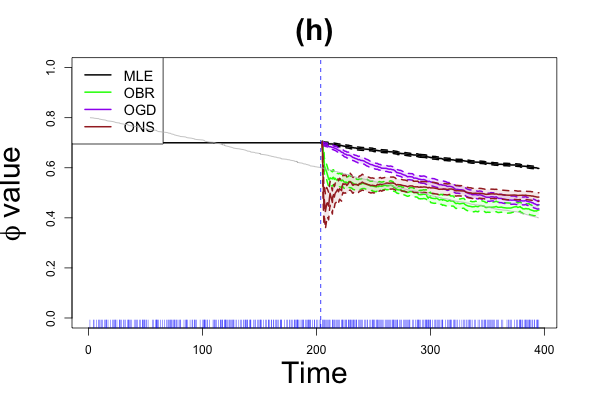}
\end{minipage}
\begin{minipage}{0.33\linewidth}
\includegraphics[width=\textwidth]{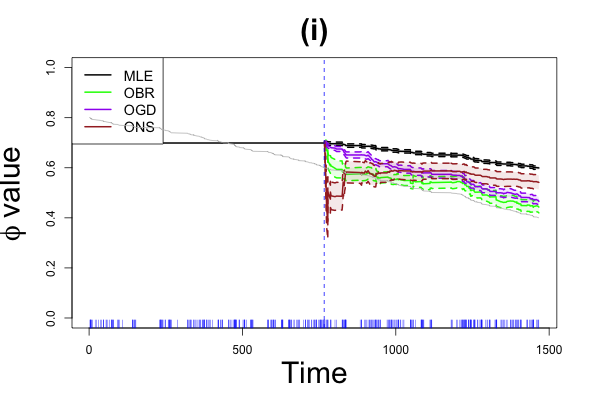}
\end{minipage}
\caption{Mean and confidence intervals of the online estimates for simulated irregular autoregressive processes. The three rows represent (from top to bottom) the sanity check, abrupt change and constant change experiments. The three columns represent (from left to right) the examples for regular, Gamma(3,3) and Exp. M(15,2,0.15,0.85) times. The black line represents the batch estimation of the iAR model. The brown, purple and green lines represent the parameter estimation using the iAR-ONS, iAR-OGD and the iAR-OBR methods respectively. Finally, the gray line is the true parameter with which the time series were simulated. \label{fig:sim}} 
\end{figure*}
\end{center}

Furthermore, the complete trajectory of the parameters estimated using the proposed methods is presented in Figure \ref{fig:sim}. Note from Figures \ref{fig:sim} (a)-(b)-(c) that the online estimation methods converge quickly to the true parameter, particularly when the time gaps are small.\\

Table \ref{sim2} shows the results of the Monte Carlo simulations for the Abrupt Change scenario in which the parameter of each simulated process of size n=400, starts with a parameter of $\phi = 0.7$ for the first 200 observations and changes to $\phi = 0.3$ for the last 200 observations. As can be noticed, the proposed online estimation methods can  adapt to the structural change produced in the time series giving a last parameter estimation closer to the true parameter of the second half of time series, unlike the batch estimation which obtains an approximated last estimated parameter of 0.5 for each observational times. The fact that the batch estimation fails to estimate well the parameters used to generate each time series impact on the time series fit, as can be seen from the mean squared error value. As in the sanity check scenario, the online method that obtains the best fit to the series is the iAR-OBR method, which achieves the lowest mean squared error for the four distributions used to generate the observational times. Figures \ref{fig:sim} (d)-(e)-(f) shows that the iAR-OGD method converges slower to the true value of the parameter than the remaining online estimation methods.\\

\begin{table}
\begin{centering}
\resizebox{\columnwidth}{!}{
\begin{tabular}{|c|c|c|c|c||c|c|c|c|}
  \hline
\textbf{Obs. Time} & \textbf{$\hat{\phi}_{400}$ MLE} & \textbf{$\hat{\phi}_{400}$ OBR} & \textbf{$\hat{\phi}_{400}$ OGD} & \textbf{$\hat{\phi}_{400}$ ONS} & \textbf{MSE MLE} & \textbf{MSE OBR} & \textbf{MSE OGD} & \textbf{MSE ONS} \\ \hline \hline
Regular & 0.502 & \textbf{0.294} & 0.309 & 0.319 & 0.93 & \textbf{0.84} & 0.94 & 0.96 \\ 
  Unif(0.5,1.5) & 0.494 & \textbf{0.299} & 0.316 & 0.31 & 0.92 & \textbf{0.82} & 0.91 & 0.92 \\ 
  Gamma(3,3) & 0.497 & 0.282 & 0.327 & \textbf{0.3} & 0.85 & \textbf{0.74} & 0.82 & 0.82 \\ 
  Exp. M(15,2,0.15,0.85) & 0.492 & \textbf{0.311} & 0.369 & 0.399 & 0.85 & \textbf{0.8} & 0.86 & 0.91 \\   
   \hline
\end{tabular}
}
\end{centering}
\caption{Summary of the Monte Carlo experiments for the Abrupt Change scenario performed using a parameter $\phi=0.7$ for the first half and $\phi=0.3$ for the second half of each simulated time series for the four distributions used to generate the observational times. Columns 2 to 5 of this table show the last value estimated for each method. The last four columns show the mean squared error computed for the fitted values. \label{sim2}}
\end{table}

The last simulation experiment that we perform is the Constant Change scenario. In this example the parameter of the iAR model decreases constantly from an initial value of $\phi = 0.8$ to a final value of $\phi = 0.4$. Table \ref{sim3} shows that, as in the abrupt change scenario, the batch estimation of the iAR model parameter is very distant from the true parameter. In particular, the last value estimated by the batch method is close to the average of the values used to generate the simulated time series. In contrast, the online estimation methods achieve values closer to the last parameter. Here the iAR-OBR method stands out with a final parameter estimate closer to 0.4. Regarding the goodness of fit, the iAR-OBR method again has the best fitted values according to the mean squared error. Finally, Figures \ref{fig:sim} (g)-(h)-(i) shows a very slow decay of the batch estimation, while the online estimations adapt quickly to the true parameter, particularly the iAR-OBR and iAR-ONS methods.

\begin{table}
\begin{centering}
\resizebox{\columnwidth}{!}{
\begin{tabular}{|c|c|c|c|c||c|c|c|c|}
  \hline
\textbf{Obs. Time} & \textbf{$\hat{\phi}_{400}$ MLE} & \textbf{$\hat{\phi}_{400}$ OBR} & \textbf{$\hat{\phi}_{400}$ OGD} & \textbf{$\hat{\phi}_{400}$ ONS} & \textbf{MSE MLE} & \textbf{MSE OBR} & \textbf{MSE OGD} & \textbf{MSE ONS} \\ \hline \hline
  Regular & 0.604 & \textbf{0.409} & 0.416 & 0.479 & 0.75 & \textbf{0.64} & 0.69 & 0.71 \\ 
  Unif(0.5,1.5) & 0.597 & \textbf{0.427} & 0.429 & 0.483 & 0.75 & \textbf{0.65} & 0.7 & 0.72 \\ 
  Gamma(3,3) & 0.597 & \textbf{0.431} & 0.453 & 0.483 & 0.68 & \textbf{0.57} & 0.6 & 0.63 \\ 
  Exp. M(15,2,0.15,0.85) & 0.599 & \textbf{0.443} & 0.465 & 0.541 & 0.76 & \textbf{0.71} & 0.74 & 0.78 \\ 

   \hline
\end{tabular}
}
\end{centering}
\caption{Summary of the Monte Carlo experiments for the Constant Change scenario performed using an initial parameter of $\phi=0.8$ which decreases to $\phi=0.4$ for the four distributions used to generate the observational times. Columns 2 to 5 of this table show the last value estimated for each method. The last four columns show the mean squared error computed  for the fitted values.\label{sim3}.} 
\end{table}

\begin{center}
\begin{figure*}
\centering
\includegraphics[width=\textwidth]{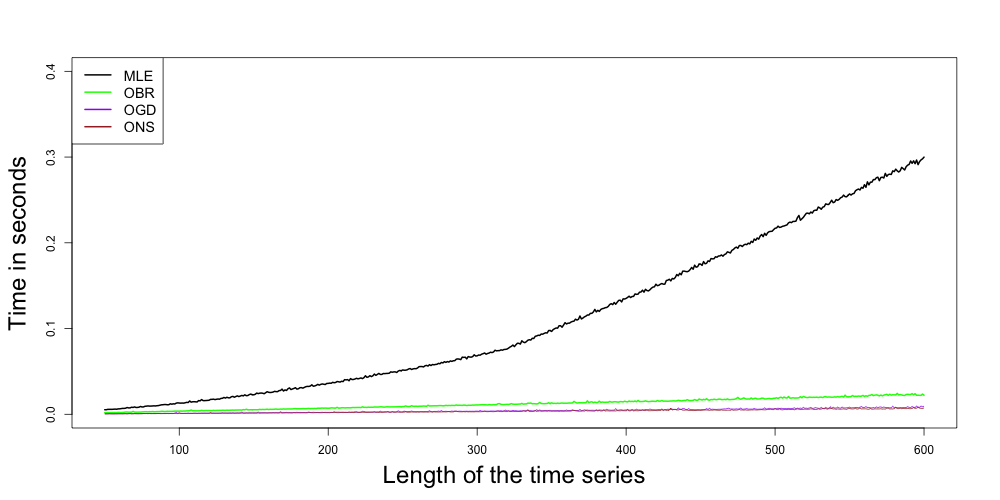}
\caption{Mean of the computation times of the estimation methods for the iAR model for simulated processes with samples sizes from n=50 to n=600. The black line represents the batch estimation of the iAR model. The brown, purple and green lines represent the parameter estimation using the iAR-ONS, iAR-OGD and the iAR-OBR methods respectively. \label{fig:time}} 
\end{figure*}
\end{center}

In order to evaluate the efficiency of the estimation methods proposed in this work, we obtained the computation times of each estimation method for different sample sizes of the simulated times series, from n=50 to n=600. For each sample size, the computation time was obtained 100 times. Later, we calculated the mean of the computation times obtained in each repetition. In Figure \ref{fig:time} can be observed that the online estimation methods are faster than the batch estimation for all the sample sizes evaluated. Particularly, for time series of 100 observations, batch estimation (0.0131 seconds) takes on average 12 times longer than iAR-OGD and iAR-ONS estimation (0.0011 seconds) and 4 times longer than iAR-OBR estimation (0.0036 seconds). Furthermore, this difference increases as more observations are required to be estimated.\\ 

\section{Application to Real-life Data}
\label{sec:application}

\subsection{Flow of the river Nile}
\label{ss:Nile}

\begin{center}
\begin{figure*}
\centering
\begin{minipage}{0.49\linewidth}
\includegraphics[width=\textwidth]{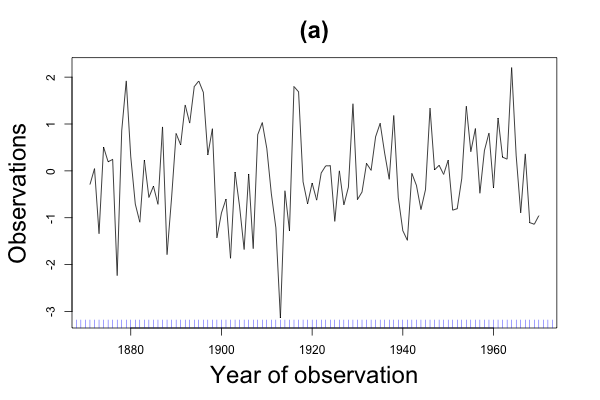}
\end{minipage}
\begin{minipage}{0.49\linewidth}
\centering
\includegraphics[width=\textwidth]{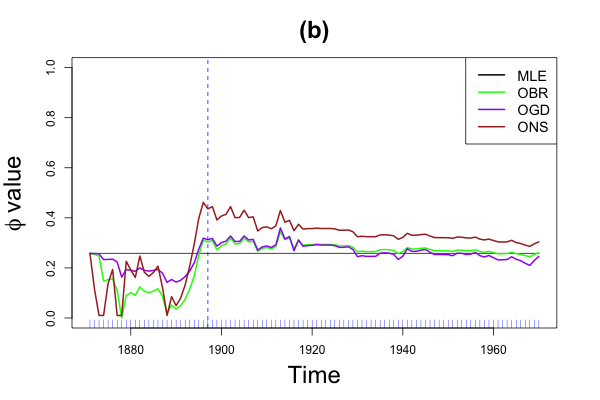}
\end{minipage}
\caption{Online estimation of the flow of the river Nile time series. Figure (a) shows the time series of the flow of the river Nile. Figure (b) shows the parameter estimation at each point of the time series. The black line represents the batch estimation of the iAR model. The brown, purple and green lines represent the parameter estimation using the iAR-ONS, iAR-OGD and the iAR-OBR methods respectively. The vertical line represents the year in which the change-point occurs. \label{Nile}}
\end{figure*}
\end{center}

For our first application we use the Nile dataset available in R. In this time series the observations corresponds to the measurement of the annual flow of the river Nile below the Aswan Dam, in $10^8$ m$^3$. This dataset contains 100 observations regularly observed between the year 1871 to 1970. The Nile time series has been analyzed in several previous studies because it has an apparent change point in the year 1898, which was first detected by by Cobb, G (1978) \cite{Cobb_78}. Figure \ref{Nile} (a) shows the Nile time series. Before implementing the estimation methods proposed in this work, we preprocessed the data to ensure a constant mean and variance equal to one. For this purpose, we first estimate the trend of the time series using the lowess method. Later, we remove the trend from the series. Finally, we standardize the de-trended time series. \\

Figure \ref{Nile} (b) shows that the online estimation methods proposed in this work detect this change point, exhibiting a break in the increasing trend of the parameter estimates observed before 1898. After this break the parameter estimates slowly decrease until reaching an estimate close to the batch estimate (0.258) at the end of the time series. Here, the iAR-OBR method has its last estimate (0.261) closest to the Batch estimate of the parameter of the iAR model.

\subsection{Infant heart rate}
\label{ss:Baby}

\begin{center}
\begin{figure*}
\centering
\includegraphics[width=\textwidth]{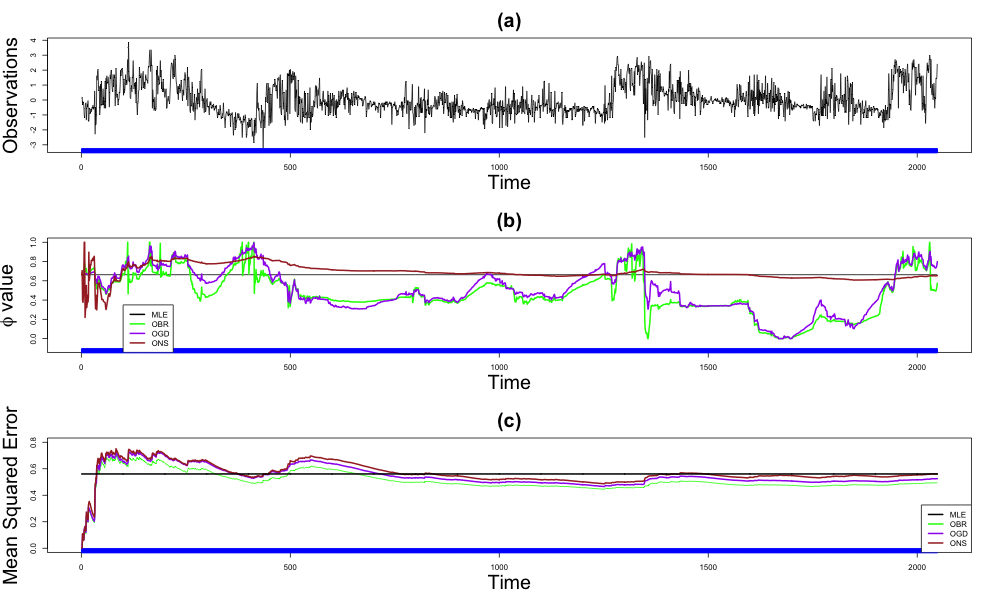}
\caption{Online estimation of the infant heart rate time series. Figure (a) shows the time series of the infant heart rate. Figure (b) shows the parameter estimation at each point of the time series. Figure (c) shows the mean squared error estimated at each point of the time series. The black line represents the batch estimation of the iAR model. The brown, purple and green lines represent the parameter estimation using the iAR-ONS, iAR-OGD and the iAR-OBR methods respectively.\label{Baby}}
\end{figure*}
\end{center}

The second real dataset that we use in this work is the time series of the record of an 66 day old infant heart rate (in beats per minute).This time series is available in the R package wavethresh \cite{wavethresh} under the name BabyECG. The time series contains 2048 observations sampled regularly every 16 seconds between the 21:17:59 of a day and the 06:27:18 of the following day. This data was introduced by Nason et al. (2000) \cite{Nason_00} as an example of a locally stationary time series due to the variations in its behavior over time, as can be seen in Figure \ref{Baby} (a). Considering these variations over time, the BabyECG time series is an interesting example to assess the ability of online estimation methods to adapt to structural changes in the time series. As in the previous example, we remove the trend from the time series and then standardize it before applying the estimation methods.\\

Figure \ref{Baby} (b)  shows a high variability in the parameter estimates obtained from the online estimation methods, particularly for the iAR-OGD and iAR-OBR. This result indicates that the online estimation methods are being able to detect the structural changes in the time series. Furthermore, this can be validated from the goodness of fit obtained for this time series from the online estimations. Note in Figure \ref{Baby} (c) that the mean squared error obtained by the online estimation methods are consistently lower than those obtained from the batch estimation for more than half of the observations in the time series. By calculating the mean of the estimated mean squared errors for each observation, we note that the method with the best fit of this time series is the iAR-OBR with an average MSE of 0.51. This value is lower than the MSE obtained from the batch estimation (0.56).

\newpage

\subsection{Brightness of astronomical object}
\label{ss:Astro}

\begin{center}
\begin{figure*}
\centering
\includegraphics[width=\textwidth]{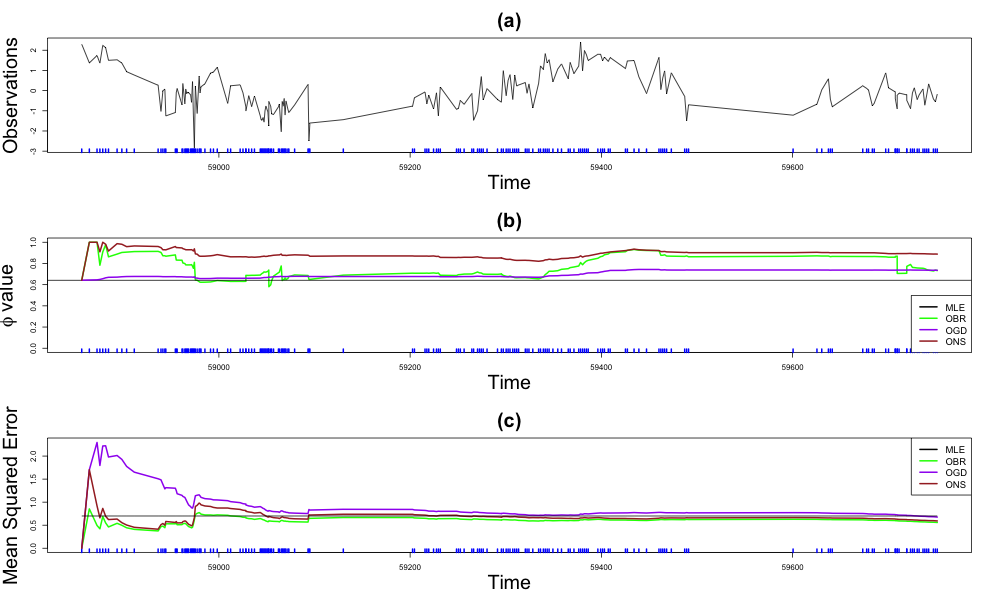}
\caption{Online estimation of the brightness of an astronomical object. Figure (a) shows the time series of the brightness of a changing state-AGN candidate Figure (b) shows the parameter estimation at each point of the time series. Figure (c) shows the mean squared error estimated at each point of the time series. The black line represents the batch estimation of the iAR model. The brown, purple and green lines represent the parameter estimation using the iAR-ONS, iAR-OGD and the iAR-OBR methods respectively.\label{AGN}}
\end{figure*}
\end{center}

For the last application to real data we use the time series of the brightness of an astronomical object. Unlike the time series used in the previous examples, this time series is irregularly observed. The astronomical object we consider in this example corresponds to an Active Galactic Nuclei (AGN) observed in the ZTF survey \cite{Bellm_2018} and coded as ``ZTF20aagiimu''. The data processing was done by the broker ALeRCE \cite{Forster_2021}. This object has been studied by Sanchez-Saez et al (2021) \cite{Sanchez_Saez_2021_2}, who have proposed it as a candidate for a Changing-state AGN, i.e. an object that changes state in time, so it is interesting to see if the estimation methods proposed in this work are able to detect these behavioral changes. This time series contains 177 observations that were measured in more than 3 years. This astronomical time series is presented in Figure \ref{AGN} (a). Before implementing the estimation methods, we did the preprocessing of the time series explained in the example \ref{ss:Nile}.\\  

Figure \ref{AGN} (b) shows that the three online estimation methods obtain an abrupt increase in the estimated parameter during the second half of the time series, which may be explained by a change of the state of this astronomical object. Furthermore, if we observe the mean squared error obtained for each estimation method we notice that again these values are lower for the iAR-OBR method (Figure \ref{AGN} (c)), which reaches an average MSE of 0.6, while the Batch estimation reaches an MSE of 0.7. According to this result, a parameter estimate which is able to be adapted to the new observations obtains better fitted values for this time series than one that uses all the observed data.

\section{Discussion}
\label{sec:discussion}

In this work we presented three online estimation methods for the irregular autoregressive (iAR) model. These methods are the first approach in the literature to the online estimation of irregularly observed autoregressive processes, since so far the online estimation methods applied to time series have been implemented on regular-time ARMA processes.\\

In the experiments carried out throughout this paper, it is observed that the proposed methods show at least two advantages over batch estimation. First, the online estimation methods have proven to be capable of adapting to structural changes in the processes. In addition, we show that the proposed methods have shorter computation times than the batch maximum likelihood estimator.\\

Among the proposed estimation methods, the iAR-OBR method achieves the best goodness-of-fit indicators in the time series analyzed.  In addition, this method showed a faster adaptation to the structural changes of the time series. On the other hand, the iAR-ONS and iAR-OGD methods are less computationally expensive.\\

Based on the results obtained, we consider that online estimation methods can make an efficient estimation of streaming data, where new data arrive frequently and models must be updated to avoid becoming obsolete. This efficiency is reach in terms of estimation accuracy and computation time.\\

In future works, we aim to extend these methods to perform online estimations of other time series models for irregularly observed data (such as the complex irregularly observed autoregressive model (CiAR \cite{Elorrieta_2019}) and the bivariate irregularly observed autoregressive model (BiAR \cite{Elorrieta_2021}) models).


\bibliographystyle{splncs03}
\bibliography{bibonline}


\end{document}